\documentclass{article}
\usepackage{main_sty}

\usepackage{times}
\usepackage{soul}
\usepackage{url}
\usepackage[hidelinks]{hyperref}
\usepackage[utf8]{inputenc}
\usepackage[small]{caption}
\usepackage{graphicx}
\usepackage{amsmath}
\usepackage{amsthm}
\usepackage{booktabs}
\usepackage{algorithm}
\usepackage{algorithmic}
\usepackage[switch]{lineno}
\usepackage{graphicx}
\usepackage{subfigure}
\usepackage{multirow}

\title{IB-Net: Initial Branch Network for \\Variable Decision in Boolean Satisfiability}

\author{
Tsz Ho Chan$^1$
\and
Wenyi Xiao$^1$\and
Junhua Huang$^1$\And
Huiling Zhen$^1$\And
Guangji Tian$^2$\And
Mingxuan Yuan$^1$\footnote{Corresponding author}
\affiliations
$^1$Noah's Ark Lab, Huawei\\
$^2$Hisilicon, Huawei\\
\emails
chantszho1@huawei.com,
xiaowenyi2@huawei.com,
huangjunhua15@huawei.com,
zhenhuiling2@huawei.com,
tianguangji@hisilicon.com,
Yuan.Mingxuan@huawei.com
}

\begin{document}

\maketitle

\begin{abstract}
Boolean Satisfiability problems are vital components in Electronic Design Automation, particularly within the Logic Equivalence Checking process. Currently, SAT solvers are employed for these problems and neural network is tried as assistance to solvers. However, as SAT problems in the LEC context are distinctive due to their predominantly unsatisfiability nature and a substantial proportion of UNSAT-core\footnote{An UNSAT-core is a unsatisfiable subset of the original clauses in a Boolean expression. Its removal would render the problem satisfiable, aiding in identifying the cause of unsatisfiability.} variables, existing neural network assistance has proven unsuccessful in this specialized domain. To tackle this challenge, we propose IB-Net, an innovative framework utilizing graph neural networks and novel graph encoding techniques to model unsatisfiable problems and interact with state-of-the-art solvers. Extensive evaluations across solvers and datasets demonstrate IB-Net's acceleration, achieving an average runtime speedup of 5.0\% on industrial data and 8.3\% on SAT competition data empirically. This breakthrough advances efficient solving in LEC workflows.\footnote{The implementation code will be provided after submission.}
\end{abstract}

\section{Introduction}
In computational theory and complexity, the Boolean Satisfiability (SAT) problem stands as a cornerstone. This decision problem, underpinned by the task of identifying the possibility of truth assignment to the Boolean expression, is known to be NP-complete and computationally demanding \cite{cook1971}. 
SAT is fundamental to various practical applications, including software verification, hardware design, and more \cite{kasi2013} \cite{DBLP:conf/lpar/Leino10}. Those problems will first be reduced into the SAT formula and then apply SAT solvers. During the development of SAT solvers, various techniques have been applied to boost the efficiency of the solving process. There are currently two major categories, Stochastic Local Search (SLS) based solvers and Conflict Driven Clause Learning (CDCL) based solvers. SLS solvers perform initial assignments for all variables in the formula and repeatedly flip the variable assignment to maximize the internal score, which leads to a solution. The foundation of CDCL solvers are clause learning and deep backtracking search, which make assignment for one variable each time until the conflict occurs to perform analysis and backtracking.

Following the ubiquity of SAT, its implication is particularly critical in Electronic Design Automation (EDA), the suite of software used in designing electronic systems and chips. Specifically, Logic Equivalence Checking (LEC), an essential step in verifying the correctness of a circuit design after transformation in the EDA process, leverages the efficiency of SAT solvers. LEC aims to determine whether two circuits are functionally equivalent by encoding the checking problem into Boolean expressions and applying the SAT solver. As this process is to detect the potential fault in transformation, which should exist with a small possibility, we would expect that most SAT problems in LEC are unsatisfiability (UNSAT): no possible variables assignment can be found. The SAT solver will be called by LEC for numerous times for each circuit design during the design verification stage. Thus the performance of applied SAT solvers would be the critical point to boost the efficiency of LEC process.

In recent years, there have been some studies utilizing neural networks (NN), mostly graph neural networks (GNN), to solve SAT problems directly or assist current SAT solvers. 
NeuroSAT pioneers the usage of NN in SAT by proposing an end-to-end model to predict the satisfiability of an SAT problem and further predict the assignment of variables~\cite{neurosat}. NLocalSAT adopts the model design of NeuroSAT and uses the model to boost SLS solvers offline, modifying the initial variable assignment with the output of NN for one time~\cite{nlocal}. 
However, it utilizes SAT/UNSAT for the instances as the supervision signal, which is invalid in the LEC verification scenario. Although Neurocore focuses on UNSAT problems, it computes the variable assignment periodically, which is not suitable for high-frequency verification in LEC~\cite{neurocore}.

Therefore, we propose the Initial Branch Network (IB-Net), a framework that targets UNSAT problems and interacts with state-of-the-art CDCL solver to perform offline one-time branch initialization and boost efficiency of end-to-end SAT-solving process. 
We design Weighted Literal-Incidence Graph to encode the Boolean formula and employ GNN to predict the possibility of UNSAT-core variables. By proposing novel graph node feature embedding and loss function, we address the imbalanced data distribution of problems in LEC (mostly UNSAT problems and large UNSAT-core). We evaluate IB-Net and previous works with CDCL solvers on both the open SAT Competition dataset and the real industrial dataset. IB-Net achieves a 5.0\% runtime reduction per problem in the whole LEC pipeline, while other benchmarks have no effect or even runtime increase. 
For the open SAT Competition dataset, IB-Net gains an 8.3\% runtime reduction, much higher than other works. Such experimental results show the efficiency-boosting of IB-Net. All runtimes are computed with network construction and inference time inclusive.

\section{Related Work}
The EDA field has leveraged the power of NN for various applications. The complexity of EDA tasks, including verification like LEC, provides a fertile ground for applying machine learning techniques to improve efficiency and robustness. In early work, these approaches focused mainly on data processing and process optimization. NN-based data argumentation and instance preprocessing are applied to accelerate the verification process~\cite{junhua}. Huang et al. further demonstrated the utility of NNs in EDA by designing a machine learning model to predict cut ranking for dealing with combinatorial optimization problems~\cite{cuts}. 

As SAT solver is widely applied in EDA, the combination of NN and SAT problem has also been tried. Graph Neural Network has emerged as a powerful tool for capturing the relational information inherent in graph structures, which has naturally led to their application in SAT constraint problem-solving~\cite{gnnsat}. The application of GNN is approached in two ways: end-to-end solving and heuristic solver guidance. 
In the first vein, GNN is trained to learn the underlying structure and patterns in SAT problems and to directly predict satisfying assignments or indicate unsatisfiability~\cite{neurosat}~\cite{Amizadeh2018LearningTS}~\cite{xusat}. While the idea of using GNN as a solver is enticing, it also presents challenges. The complex nature of SAT problems leads to unreliability and low performance in models, especially in the industrial context. On the other hand, the use of GNN as heuristics to SAT solvers constitutes another effective application. In these approaches, GNN is typically employed to predict key aspects of the solving process, such as variable importance or ranking, thereby guiding the solver to solve efficiently~\cite{neurocore}~\cite{nlocal}~\cite{gnnli}~\cite{10.5555/3327546.3327694}. Unlike the first approach, these methods just provide information that potentially accelerates the solving process while the solving process is still done by the solver. Thus the reliability and complexity of SAT problem solving is ensured. In light of existing methodologies, our work IB-Net, a model designed to enhance the robustness and efficiency of SAT solvers through heuristic guidance and explore solver-GNN interaction, aims to address SAT solver usage in specific industrial scenarios.

\section{IB-Net approach}
\subsection{Overview}

\begin{figure*}[t] 
\centering 
\includegraphics[width=0.85\textwidth]{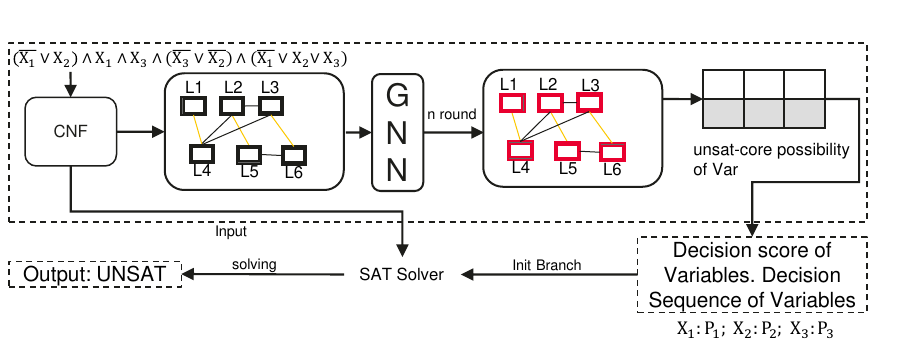} 
\caption{Model Overview} 
\label{Fig.model} 

\end{figure*}

To address our goal to assist the CDCL solvers targeting UNSAT problems, we propose the framework in Fig. \ref{Fig.model}. It is a combination of a neural network and a method to interact with solvers. For an input Boolean formula, it will firstly be converted into a Weighted Literal-Incidence Graph (WLIG) and fed into GNN for node feature updating, then gather the output as possibility to be UNSAT-core for variables. These output are assigned to SAT CDCL solver to initialize variable decision queue and decision score to guide the solving process. To make IB-Net more suitable for dealing with UNSAT problem, we made serveral improvement along the pipeline, which will be discussed in following subsections.

\subsection{Graph Formulation}

A propositional logic formula is a Boolean expression that contains conjunctions, disjunctions, variables, negations, and constants. The Boolean formula can be expressed in conjunctive normal form (CNF) with conjunction of disjunctions of variables (possibly negated variables) by transformation with linear length in linear time~\cite{cnf}. The SAT instance in CNF is a conjunction of clauses while a clause is a disjunction of literals (the variables and negated variables). The SAT instances are already converted into CNF before accessed by model and solvers, which is a common step when apply solvers. 

To make use of the structure information within the formula, we convert SAT instance \textbf{S} into WLIG \textbf{G}.  But in WLIG, the literal set of \textbf{S} are nodes and the common incidence of literals are edges, i.e. ($l_1$, $l_2$) hold edge if $l_1$ and $l_2$ appear in the same clause and the weight on edge account for the number of common incidences. The adjacent matrix $A$ of graph \textbf{G} will be the input to the graph model. 

\subsection{Neural Network Model}

We use a weighted GCN (WGCN) \cite{wgcn} as our first network to extract features of variables and a multi-layer perceptron (MLP) to output the possibility of variables presented in UNSAT-core. The WGCN can perfectly match the design of WLIG to utilize the edge weight and enhance the message exchange.

The WGCN accepts the adjacency matrix as input to build the graph structure. Apart from that, to address features of literals, we initialize the embedding for a literal node with vector $h$ $\in$ \textbf{R}$^{2d}$. The vector $h$ is produced by concatenating the degree of node $D$ $\in$ \textbf{R} and the literal type of node $T$ $\in$ \textbf{R} vertically and then feeding to a linear transformation $L_{init}$: \textbf{R}$^{1\times2}$ $\rightarrow$ \textbf{R}$^{1\times2d}$ to produce sparse node embedding. 
\begin{equation}
    h_i = L_{init}(D_i \oplus T_i), \
    H = [h_1, h_2, \ldots, h_N]^T
\end{equation}
The node embedding vectors $H$ will be updated by aggregating embedding from its neighbors during each iteration of WGCN. Namely, a single iteration can be represented: 
\begin{equation}
    H^{(l+1)} = \text{{ReLU}}({L_{out}}(A'H^{(l)} \oplus \text{{Flip}}(H^{(l)}))),
\end{equation}
where $H^{(l)}$ represents the node features after the $l$-th iteration.
We detail the computation process inside the WGCN unit as follows.

$A'$ is the normalized adjacency matrix which encapsulates aggregation operations, and values in $A'$ also carry weights to apply in the WGCN.
$A'$ is computed as follows to avoid gradient explosion:
\begin{equation}
A' = \frac{A}{\sum_{i=1}^{N}\sum_{j=1}^{N} A_{ij}} .
\end{equation}
Function $Flip$ and linear transformation $L_{out}: \textbf{R}^{1\times 4d} \rightarrow \textbf{R}^{1\times 2d}$ concatenate features of literal nodes with the corresponding negated literal nodes, forming a comprehensive representation of variables.
\begin{equation}
    \text{{Flip}}(H)= [h_{\frac{N}{2}+1}, \ldots, h_N, h_1, \ldots, h_{\frac{N}{2}}]^T
\end{equation}
After $L$ iterations of WGCN, we obtain the vectors representing structural information of literals, and feed it to MLP for the UNSAT-core possibility of each variable: 
\begin{equation}
    p_i = \text{{softmax}}(L_2 \cdot \text{{ReLU}}(L_1 \cdot  (h_i \oplus h_{N+1-i} ) + b_1) + b_2)
\end{equation}

\subsection{Training}

As we regard the UNSAT-core prediction as a classification task, the Cross-Entropy loss should be originally applied. However, due to the potential imbalance distribution (the large UNSAT-core in LEC UNSAT Problem), we utilize Focal loss~\cite{focal} as the loss function. The major difference between Focal loss and traditional Cross-Entropy loss would be the additional parameters $\alpha \& \beta$ to balance the importance of the class with a small portion. For $p$ stands for prediction possibility and $y$ is truth, the loss is computed as:

\begin{equation}
\begin{aligned}
    FL =&\sum_{i=1}^n( y_i (-\alpha (1 - p_i)^\gamma \log(p_i)) \\
    &+(1-y_i) (-(1-\alpha) p_i^\gamma \log(1 - p_i))),
\end{aligned}
\end{equation}
where $n$ is the number of variables in the target CNF.

\subsection{Interaction with Solver}
In this section, we first introduce the CDCL process and the crucial role of variable decision within it. Then we show how our model targets these stages and interacts with the process.
\subsubsection{CDCL WorkFlow}
The Conflict-Driven Clause Learning algorithm, a prevailing methodology used by modern SAT solvers, functions through a combination of decision-making, unit propagation, conflict analysis, and non-chronological backtracking. 
\begin{figure}[t] 
\centering 
\includegraphics[width=0.48\textwidth]{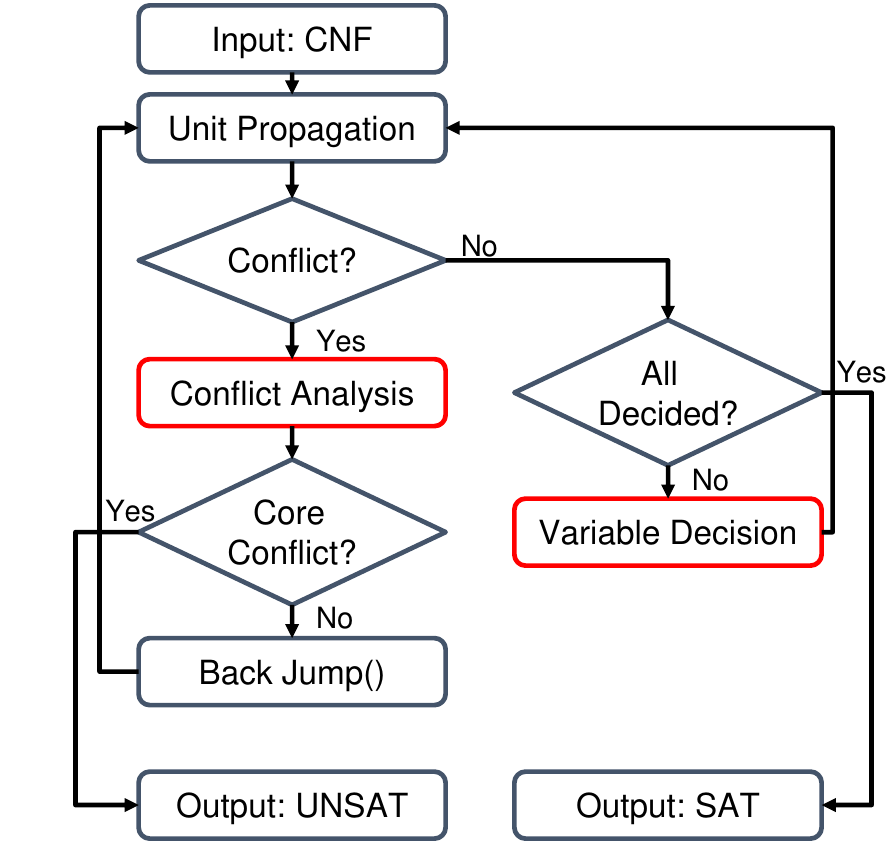} 
\caption{CDCL Process} 
\label{Fig.cdcl} 

\end{figure}

Fig. \ref{Fig.cdcl} provides a detailed workflow for the CDCL process. 

From Fig \ref{Fig.cdcl}, it can be found that 
the algorithm begins with a decision-making process where Boolean values are assigned to variables according to a decision queue. This is succeeded by unit propagation, whereby necessary assignments of other variables are deduced based on the current partial assignment. Conflicts, however, can occur if a clause in the problem becomes unsatisfiable with these assignments. In response to conflicts, the CDCL flow initiates a conflict analysis, learning and adding a new clause representing the conflict reason to the problem to avoid repetition of the same error in future iterations. The CDCL algorithm concludes with non-chronological backtracking/backjumping, which does not merely revert the most recent decision but backjumps to a point preceding the conflict, result in assigning a lower place in the decision-making queue or a lower decision-score to these variables involved in conflict. This allows a more rapid exploration of different solution spaces and aids in the avoidance of past mistakes. Two critical stages of this process are conflict analysis (clause learning) and variable decision, which collaboratively construct the decision queue for variable branching. Ideally, a swift identification of variables within the UNSAT-core would expedite the CDCL solver's resolution of UNSAT problems. Conversely, prematurely including variables outside UNSAT-core within the decision queue could slow down the solver. Thus we should try to guide the decision queue to optimize the performance. Specifically, we would instruct the decision queue and decision score, the internal variable used within variable decision and conflict analysis.

\subsubsection{Queue \& Score Based Branching}
Against above backdrop, we propose a novel methodology for interaction with the CDCL solver. Before attempting to solve a SAT problem, our trained network undertakes an inferential analysis of the problem, thereby calculating the probability of each variable being within the UNSAT-core. These probabilities guide the solver in establishing a queue as well as a set of scores for solver branch decisions. Probabilities output are first used as the initialization for decision scores. Then they are employed to rank variables in descending queue, which is adopted as the initial value for decision queue. As predicted decision queue and decision score are based on the probability of variables within UNSAT-core, solver can deal with those variables that highly related to UNSAT-core first to address the unsatisfiability.

These decision values and decision queue, now based on our neural network inference, are imported as the solver's initial values. The solver then can commence its operations on problem. This importing process doesn't undermine the solver's capacity to comprehensively address UNSAT problems, as the basic algorithm in place still centers around identifying the UNSAT-core. 

The added advantage here is that our inference performs only once for each problem, thus keeping the computational burden and associated time minimal. The processing time of the whole pipeline of IB-Net (including graph formulation, inference and solver initialization) can be controled within 1.5 second, which is regarded as a small cost comparing to the solving time. It's worth noting that our neural network is efficient enough to run on CPUs. Consequently, our method maintains the solver's completeness while significantly enhancing its efficiency and speed.

\section{Experimental Settings}

\subsection{Datasets}
\begin{table}[t]
\centering

\begin{tabular}{cc|ccc}
\toprule
\multicolumn{2}{c|}{Data}                                                               & \#CNF  & Avg\#Var & Avg\#Clause \\
\midrule
\multirow{4}{*}{LEC}                                                       & Circuit 1 & 22,050 & 1,002    & 3,970        \\
                                                                           & Circuit 2 & 14,840 & 952      & 3,664        \\
                                                                           & Circuit 3 & 25,591 & 1,563    & 6,172        \\
                                                                           & All       & 62,481 & 1,220    & 4,799        \\
\hline
\multirow{4}{*}{\begin{tabular}[c]{@{}c@{}}SAT\\ Comp\end{tabular}} & SAT       & 1,115   & 3,065     & 74,169       \\
                                                                           & UNSAT     & 985    & 3,784     & 73,751       \\
                                                                           & Halted   &1,171 & 3,623 &161,140 \\
                                                                           & All       & 3,271   & 3,481     & 105,178  \\
\bottomrule
\end{tabular}

\caption{Detail statistic of LEC circuit dataset and SAT Competition dataset. Noted that over 99\% of instance in LEC dataset is result in UNSAT.}
\label{table:data}

\end{table}
We utilize two datasets to train and evaluate our framework and previous approaches. The first one is the industrial LEC circuit SAT instances from real-life chip design in 2023. They are extracted from the real chip development process to reflect the need for LEC process. To fit the data into GPU easily and model the real-life production constraint, we first run state-of-the-art SAT solver (Kissat~\cite{kissat}) on these problems and filter the instance that solver can solve within 1,000 seconds (the instances that solver cannot solve within 1,000 seconds will be regarded as hard cases and sent to redesign in real-life). Apart from the industrial dataset, we also prepare the SAT Competition dataset as previous works are mostly target open datasets. We adopt the anniversary track of SAT Competition in 2022~\cite{sat2022}. The anniversary track is comprised of all benchmark instances used in previous SAT Competitions to ensure the coverage of problems. We also filter the anniversary track dataset in the same way with LEC dataset. The details of filtered datasets can be found in Table \ref{table:data}. $Halted$ here means instances unsolved within given time.
After preparing the datasets, we use Kissat and Drat-trim~\cite{drat} to find variable assignments for SAT problems and 
UNSAT-core for UNSAT problems, which serves as the supervision of our model training. When Kissat try to solve a UNSAT instance, it will produce a proof for the unsatisfiability, which will be taken by Drat-trim to produce the UNSAT-core variable. Though the UNSAT-core found by Drat-trim is not the minimal core, it still helps in solving UNSAT instances, so we will take these UNSAT-core as all the UNSAT-core refered in following sections. The supervision production just cost the same time of solving given dataset by Kissat, which can consider as easy to follow and efficient to produce.

\subsection{Experimental Setup}
Our experiment design commences by segregating the two previously mentioned datasets into distinct training and testing subsets, meticulously avoiding any overlaps. We allocate 80\% of the data to training, reserving the remaining 20\% for testing. 
We benchmark against several baselines: (1) The NeuroSAT~\cite{neurosat} model based on GNN and Literal-clause graph. We follow the official implementation and used the score they set for variable as our proposed probality. (2) Modified version of NeuroSAT: NeuroCore~\cite{neurocore}, that target UNSAT problem. We follow the official implementation and interaction periodically with solvers. (3) NLocalSAT~\cite{nlocal} is based on NeuroSAT and interacts with SLS solver. We adopted the official implementation but changed the output to UNSAT-core prediction. Overall, we adopt the setting for each model according to their official codes or publications but change the target output to UNSAT-core prediction. 

Our choice of the Kissat SAT solver, a top-performing Conflict-Driven Clause Learning based solver, as the collaborating solver, is guided by the comparative analysis with other CDCL solvers. From Table \ref{table:solvers}, the superior performance exhibited by Kissat in comparison with other CDCL peers justifies the selection for our study\cite{Biere2019CADICALAT}\cite{Srensson2005MiniSatV}\cite{audemard:hal-03299473}. As we just maintain an interaction with Kissat, instead of recompiling the Kissat solver, our work can be transfered to future top-performing solvers.
\begin{table}[t]
\centering
\begin{tabular}{c|cc}

\hline
    Original Solvers              & LEC         & SAT Comp    \\   \hline

CaDiCaL           & 350s         & 209s       \\     
MiniSAT           & 810s         & 450s        \\    
Glucose           & 529s         & 308s      \\      
Kissat            & 335s         & 195s    \\         
\hline
\end{tabular}
  
    \caption{Average running time of different CDCL solvers targeting random 100-sample UNSAT problems from LEC dataset and SAT Comp dataset}
  
    \label{table:solvers}
\end{table}

\subsection{Evaluation metrics}
For evaluation metrics, we aim to evaluate: (1) the efficacy of UNSAT-core prediction for UNSAT problems on test datasets. (2) end-to-end time efficiency of models in interaction with Kissat solver, compared with the standalone performance of the original Kissat solver.
We use evaluation metrics as: (1) A.RT means the average runtime among all CNFs. (2) Imp. means the efficiency improvement (reduction in seconds) compared with the original Kissat solver. (3) Halted represents the number of halted (time-out) CNFs within the given time (1,000 seconds). Please note that all the runtimes for NN-based solutions are network construction and inference time inclusive.

\section{Experimental Results}
\subsection{Main Results}

\subsubsection{UNSAT-core Prediction}

\begin{table}[t]
    \small
    \centering
    \setlength{\tabcolsep}{4pt} 
\begin{tabular}{c|ccc|ccc}
\hline
  & \multicolumn{3}{c|}{LEC Circuit} & \multicolumn{3}{c}{SAT Comp} \\ \cline{2-7} 
Approach          & Acc      & Pos.F1    & Neg.F1    & Acc    & Pos.F1   & Neg.F1   \\ \hline
NeuroCore & 89\%     & 94\%      & 27\%      & 85\%   & 77\%     & 88\%      \\
NeuroSAT  & 90\%     & 93\%      & 8\%       & 77\%   & 62\%     & 82\%      \\
NLocalSAT & 89\%     & 94\%      & 11\%       & 74\%   & 60\%     & 80\%      \\
IB-Net     & 95\%     & 97\%    & 71\%    & 91\%   & 84\%     & 93\%      \\ \hline
\end{tabular}

    \caption{Model UNSAT-core prediction performance on LEC dataset and SAT Competition dataset}
   
    \label{table:stat_label}

\end{table}

\begin{figure}[t]
\centering
\subfigure[Percentage of UNSAT-core in samples with different variable sizes for LEC dataset]{
\begin{minipage}[t]{0.4\textwidth}
\centering
\includegraphics[width=1\textwidth]{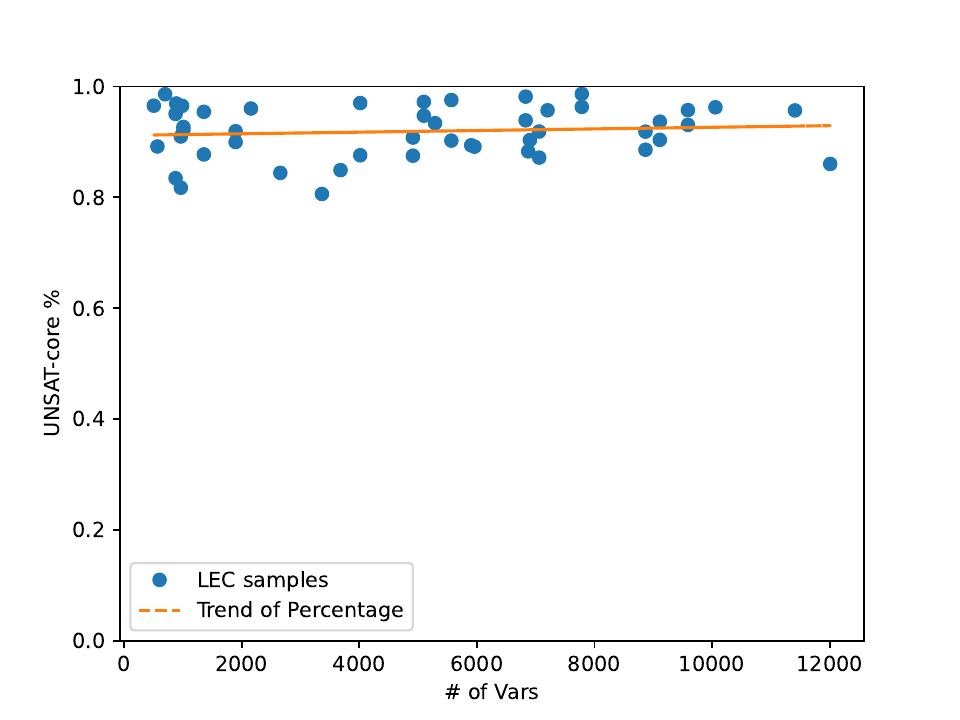}

\label{unsat_LEC}
\end{minipage}}
\subfigure[Percentage of UNSAT-core in samples with different variable sizes for SAT Comp dataset]{
\begin{minipage}[t]{0.4\textwidth}
\centering
\includegraphics[width=1\textwidth]{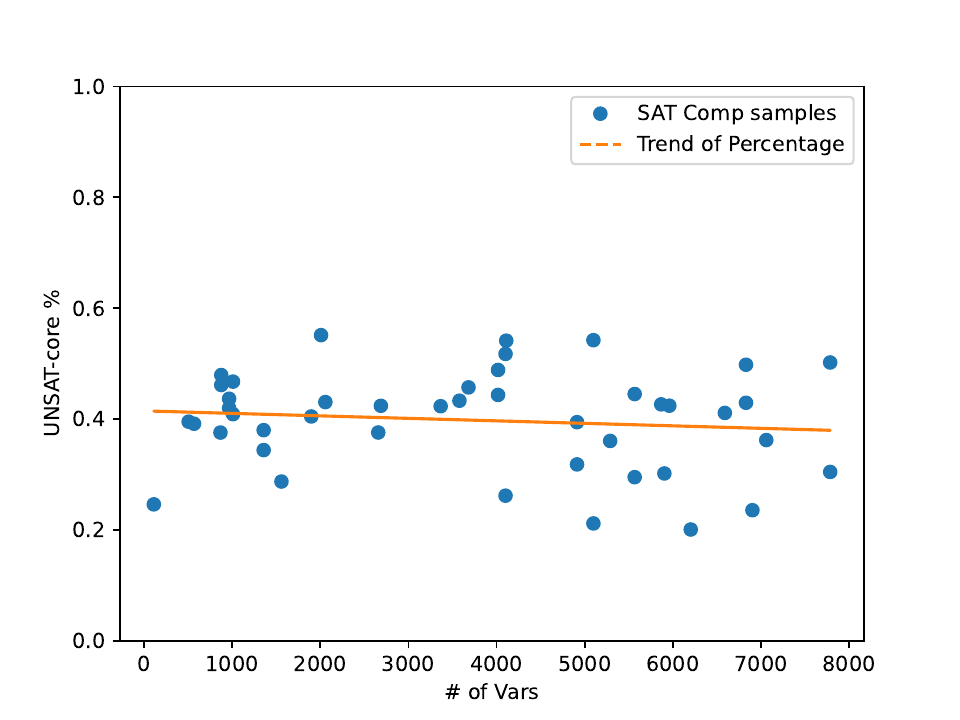}

\label{unsat_SAT}
\end{minipage}}
\caption{UNSAT-core Percentage}
\label{unsat_p}
\end{figure}

In Table \ref{table:stat_label}, we focus on the outcomes of UNSAT-core prediction. It showcases the performance of IB-Net, in comparison with other methodologies on the LEC dataset and SAT Competition dataset, which indicates that IB-Net leads the pack across both datasets. Remarkably, both F1 scores on the LEC dataset suggest IB-Net correctly identifies variables in and outside UNSAT-core. Though alternative approaches exhibit reasonable performance in SAT Competition dataset, they significantly lag behind IB-Net in LEC dataset, signifying their struggles with predicting variables in UNSAT-core accurately. This may be due to the imbalance nature of UNSAT-core in LEC UNSAT problems showed in Fig. \ref{unsat_p}: Over 90\% of variables are in UNSAT-core for LEC in Fig. \ref{unsat_LEC} while around 40\% of variables are in UNSAT-core for UNSAT problems in SAT Competition in Fig. \ref{unsat_SAT}. These results underscore the efficiency of IB-Net as a potent model for UNSAT-core prediction, with its adeptness in both balance and imbalance datasets.

\subsubsection{Runtime Reduction Performance}

\begin{table}[t]
\resizebox{\columnwidth}{!}{%
\begin{tabular}{c|c|cccc|c}
\toprule
                                 &           & Kissat & NeuroCore & NeuroSAT & NLocalSAT & IB-Net \\
\midrule
\multirow{5}{*}{LEC}     & A.RT      & 383s   & 460s      & 673s     & 680s      & 364s  \\
                                 & Imp.      & 0s     & -73s      & -285s    & -292s     & 19s  \\
                                 & Imp. \%   & 0\%      & -18.8\%   & -73\%    & -75\%     & 5\% \\
                                 & Halted    & 0      & 350       & 7800     & 7900      & 0     \\
                                 & Halted \% & 0\%      & 0.50\%     & 10\%     & 10\%      & 0\%     \\ \hline
\multirow{5}{*}{\begin{tabular}[c]{@{}c@{}}SAT \\ Comp\end{tabular}} & A.RT      & 179s   & 174s      & 176s     & 174s      & 164s  \\
                                 & Imp.      & 0s     & 5s        & 3s       & 5s        & 15s   \\
                                 & Imp. \%   & 0\%      & 3\%       & 2\%      & 3\%       & 8.3\% \\
                                 & Halted    & 1161   & 1150      & 1161     & 1158      & 1130  \\
                                 & Halted \% & 35.5\% & 35.2\%    & 35.5\%   & 35.5\%    & 34\%  \\
\bottomrule
\end{tabular}
}
\caption{Model performance on LEC dataset and SAT Competition dataset}
\label{table:stat_time}
\end{table}

\begin{table}[t]
\centering
\resizebox{\columnwidth}{!}{%
\begin{tabular}{c|c|cccc|c}
\toprule
                           &  & Kissat & NeuroCore & NeuroSAT & NLocalSAT & IB-Net \\
\midrule
\multirow{3}{*}{Circuit1} & A.RT     & 442s   & 502s      & 683s     & 685s      & 422s  \\
                           & Imp.     & 0s      & -60s      & -241s    & -243s     & 20s   \\
                           & Imp. \%  & 0\%    & -13\%     & -54\%    & -55\%     & 4.5\%   \\
\hline
\multirow{3}{*}{Circuit2} & A.RT     & 335s   & 419s      & 635s     & 640s      & 321s  \\
                           & Imp.     & 0s     & -84s      & -300s    & -305s     & 14s   \\
                           & Imp. \%  & 0\%    & -25\%     & -89\%    & -91\%     & 4.2\%   \\
\hline
\multirow{3}{*}{Circuit3} & A.RT     & 360s   & 425s      & 674s     & 695s      & 341s  \\
                           & Imp.     & 0s     & -65s      & -314s    & -334s     & 19s   \\
                           & Imp. \%  & 0\%    & -18\%     & -87\%    & -92\%     & 5.2\% \\
\bottomrule
\end{tabular}
}
\caption{Model runtime performance on three main circuits in LEC dataset. }

\label{table:case_time}

\end{table}

\begin{figure*}[ht]
\centering
\subfigure[Performance of different graph construction strategies]{
\begin{minipage}[t]{0.32\textwidth}
\centering
\includegraphics[width=5.6cm,height=3.8cm]{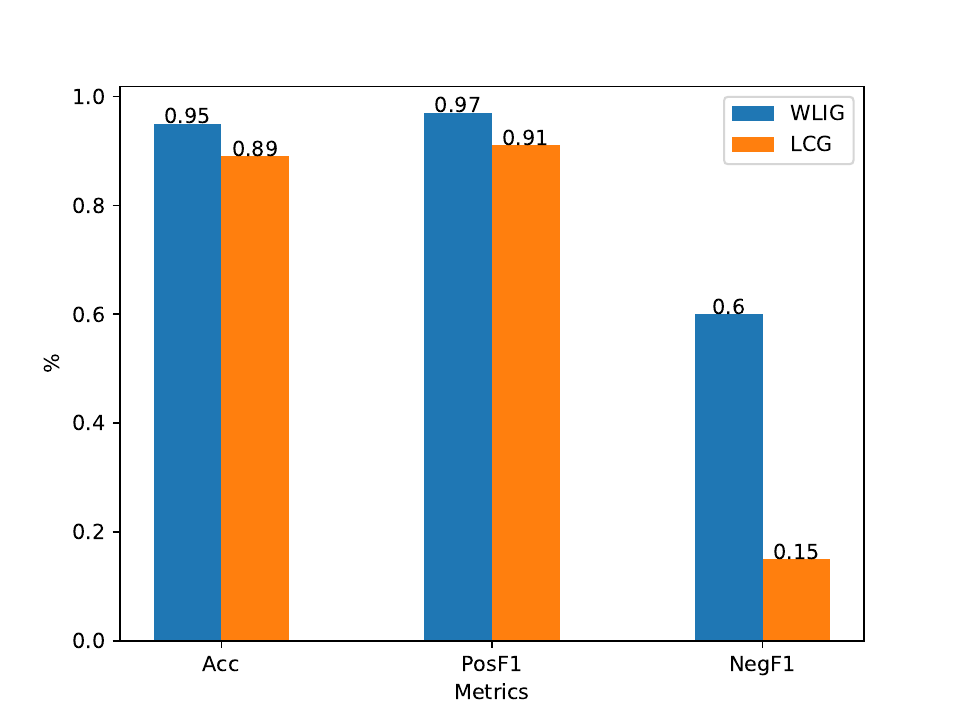}
\label{ablation_fig1a}
\end{minipage}}
\subfigure[Performance of different supervising signals]{
\begin{minipage}[t]{0.32\textwidth}
\centering
\includegraphics[width=5.6cm,height=3.8cm]{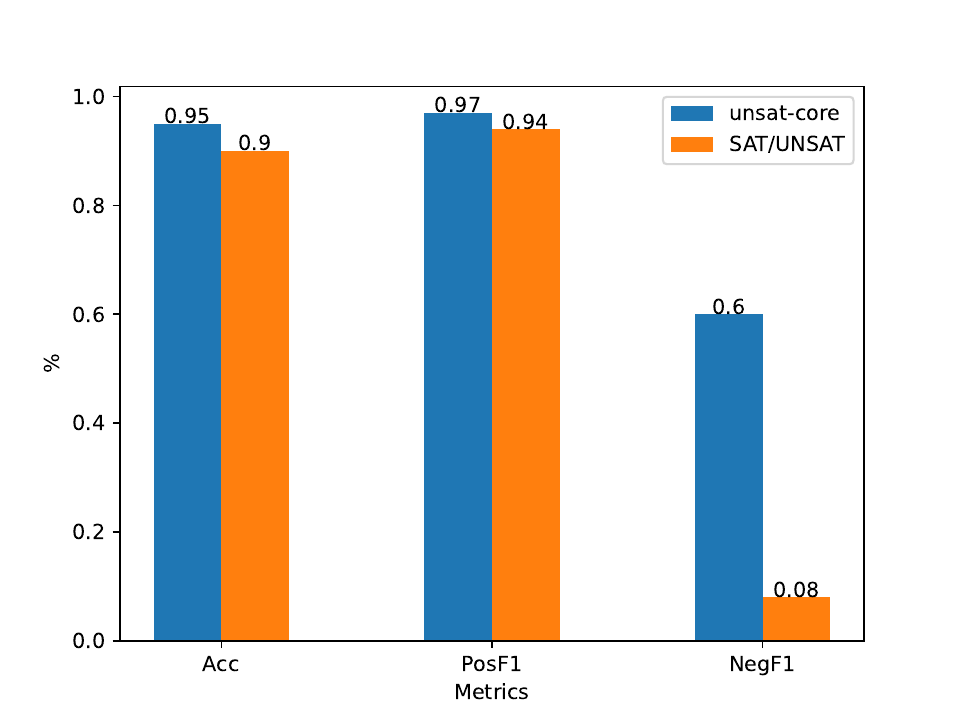}
\label{ablation_fig1b}
\end{minipage}}
\subfigure[Performance of different Loss functions]{
\begin{minipage}[t]{0.32\textwidth}
\centering
\includegraphics[width=5.6cm,height=3.8cm]{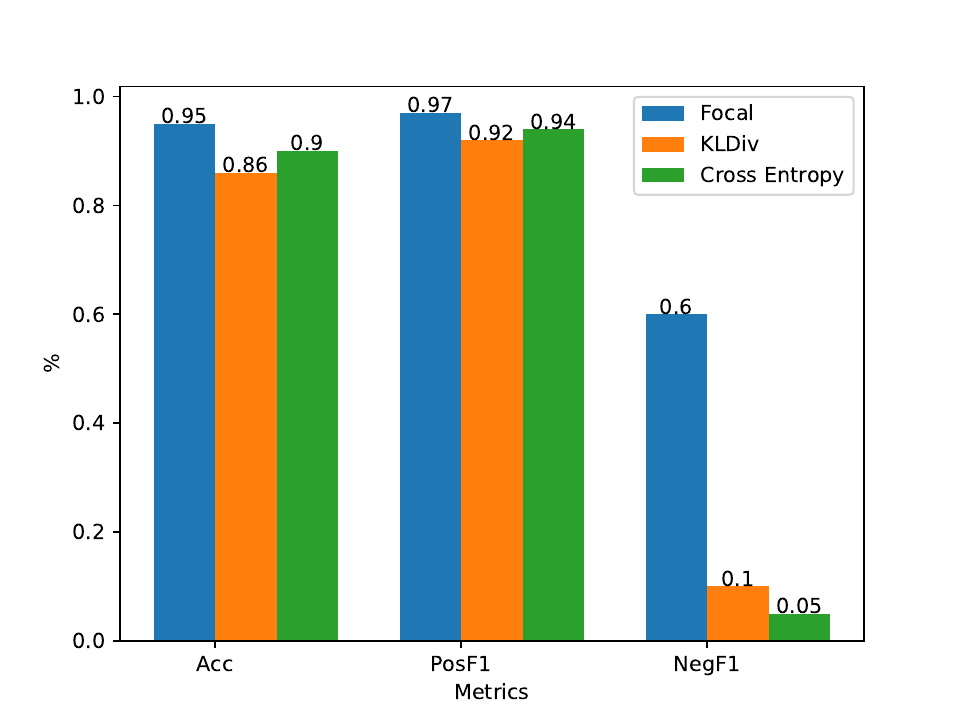}
\label{ablation_fig1c}
\end{minipage}}

\caption{The ablation study}
\label{ablation_fig1}

\end{figure*}

\begin{figure*}[ht]
\centering
\subfigure[LEC Circuit 1]{
\begin{minipage}[t]{0.32\textwidth}
\centering
\includegraphics[width=5.6cm,height=3.8cm]{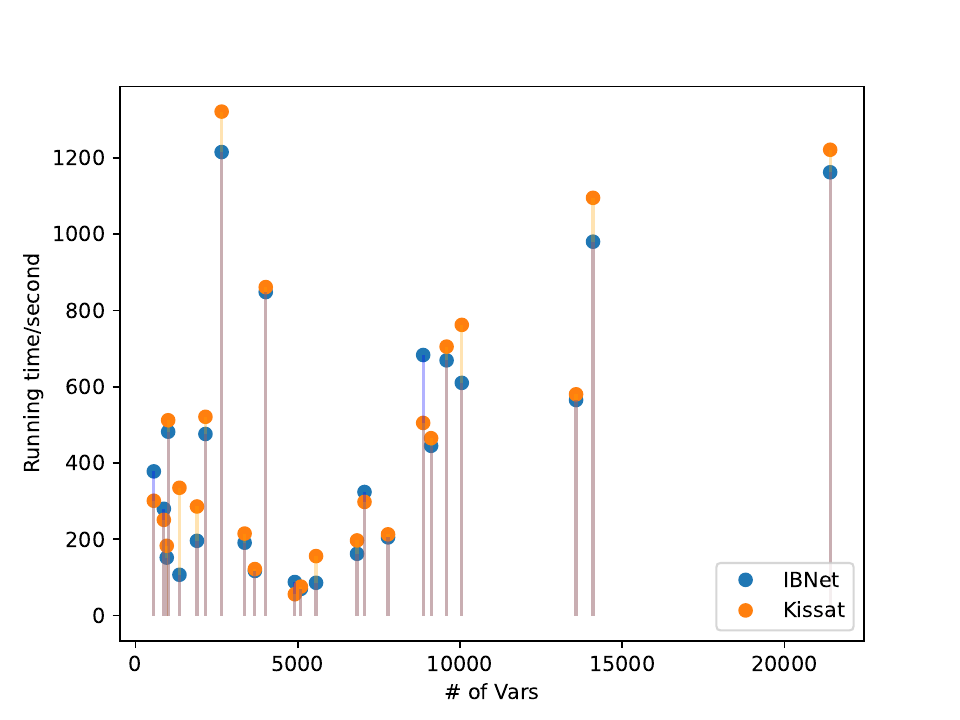}
\label{casea}
\end{minipage}}
\subfigure[LEC Circuit 2]{
\begin{minipage}[t]{0.32\textwidth}
\centering
\includegraphics[width=5.6cm,height=3.8cm]{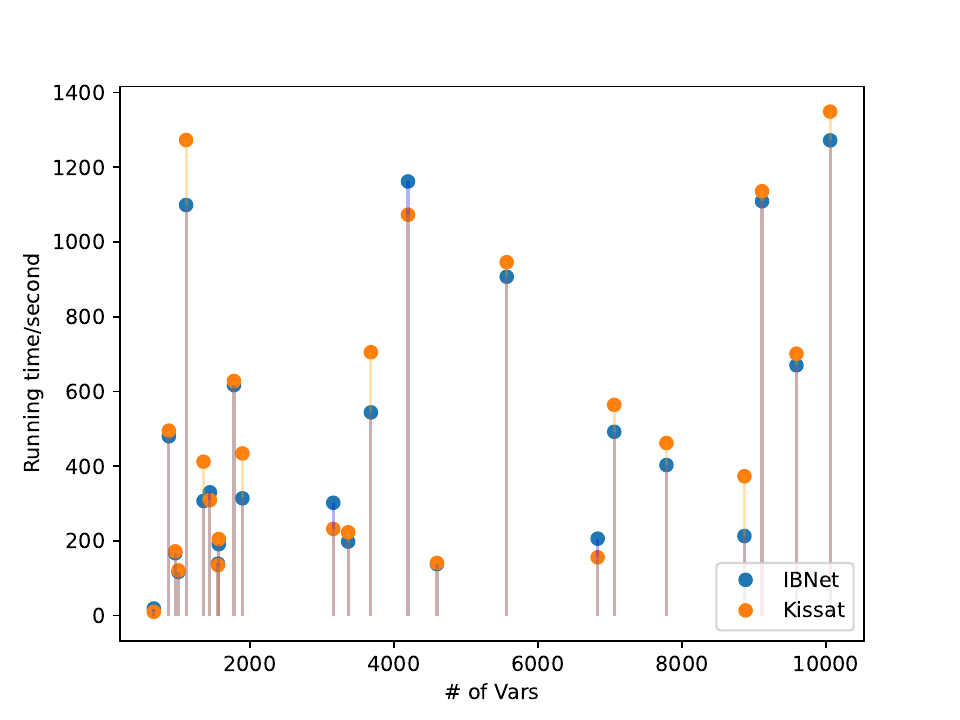}
\label{caseb}
\end{minipage}}
\subfigure[LEC Circuit 3]{
\begin{minipage}[t]{0.32\textwidth}
\centering
\includegraphics[width=5.6cm,height=3.8cm]{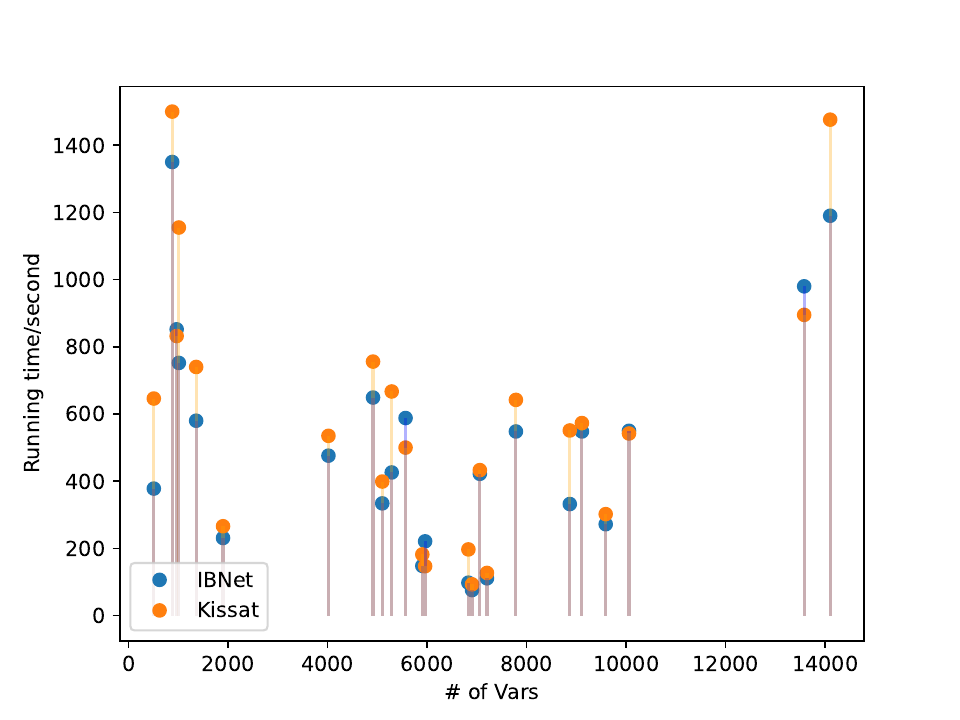}
\label{casec}
\end{minipage}}
\caption{The circuit study}
\label{case_fig1}
\end{figure*}

Table \ref{table:stat_time} provides a comparative analysis of end-to-end runtime performance for various approaches. The original Kissat is set to be the baseline. IB-Net achieves the shortest average runtime and shows improvement over the benchmark Kissat on both datasets. Specifically, IB-Net improves the average runtime by 19 seconds, accounting for 5\%, and 15 seconds, accounting for 8.3\%, on the LEC Circuit and SAT Competition respectively. Other methods exhibit much longer average runtime compared to Kissat and face additional halted instances while IB-Net can further reduce halted instances.

As we mainly target the UNSAT problem in LEC, we present Table \ref{table:case_time} to explore the detailed performance of different approaches on three subsets that constitute a significant proportion of the LEC industrial dataset. It can be observed that across these subsets, IB-Net consistently leads the board, even outperforming its improvements on the overall LEC data, which validates the position of IB-Net as a universally effective approach.

From Table \ref{table:stat_label} and Table \ref{table:stat_time}, we can observe a consistency between the quality of UNSAT-core prediction and the improvement in runtime efficiency when dealing with UNSAT problems.
It is plausible that the solver spends more time eliminating variables outside UNSAT-core from the decision queue, thus negatively impacting the performance. By fine-tuning the UNSAT dataset, our approach promises to offer more precise and time-efficient solution for SAT problems.

\subsection{Ablation Study}

We run experiments on LEC datasets with certain settings changed while other settings kept to further assess factors that may influence the effectiveness of our method on SAT problems in LEC context, shown in Fig. \ref{ablation_fig1}.

\subsubsection{Graph Construction}
In previous approaches, Literal-Clause Graph construction is applied as the common way to set both literal and clause as node and affiliation between clause and literal as edge.
The change from LCG to WLIG give up the useless clause information in UNSAT-core prediction and focus more on literal connection.

Through experiments in Fig. \ref{ablation_fig1a}, we conlcude that the WLIG construction consistently outperforms Literal-Clause Graph commonly used in previous approaches, in terms of both accuracy and F1 scores targeting UNSAT problems and especially increase the performance on variables not in UNSAT-core. This evidence supports our decision to adopt WLIG for graph construction in our model.

\subsubsection{Supervision Signals}
Results in Fig. \ref{ablation_fig1b} indicate that the UNSAT-core prediction as objective outperforms the SAT/UNSAT prediction when targeting industrial LEC dataset. The improved performance underscores the validity of our choice to employ UNSAT-core prediction as supervision signal in our model.

\subsubsection{Loss Functions}
To target the special UNSAT problem in LEC context, we amend the Cross-Entropy loss into focal loss.
Upon analysis in Fig. \ref{ablation_fig1c}, we find that the Focal loss delivers superior performance, outpacing both KLDiv and Cross-Entropy used in previsous approaches and considered common choices in classification. This result justifies our adoption of Focal loss as the preferred loss function in the imbalance data distribution.

\subsection{Circuit Study}

To further validate the effectiveness of our proposed method, IB-Net, we conducted a comprehensive case study focusing on three circuits that constitute a significant proportion of the LEC industrial dataset. For each of these circuits, we compared the runtime of IB-Net with that of the original Kissat method. Scatter plots in Fig. \ref{case_fig1} present the comparisons. In these plots, each point corresponds to a single SAT problem from one of the three targeted circuits.

The plots illustrate a clear trend: across a wide range of problem sizes, IB-Net consistently outperforms Kissat. This performance advantage of IB-Net holds true when facing substantial variability in size of problems within these circuits, which indicate that our method is not only faster but also scalable and capable of handling large and complex problems efficiently. These findings underscore the potential of IB-Net to improve the performance of SAT solver in LEC, confirming the practical applicability of our method in a real-world setting.

\subsection{Scalability}
\begin{figure}[t]
\centering
\subfigure[Memory Cost of different neural network]{
\begin{minipage}[t]{0.43\textwidth}
\centering
\includegraphics[width=1\textwidth]{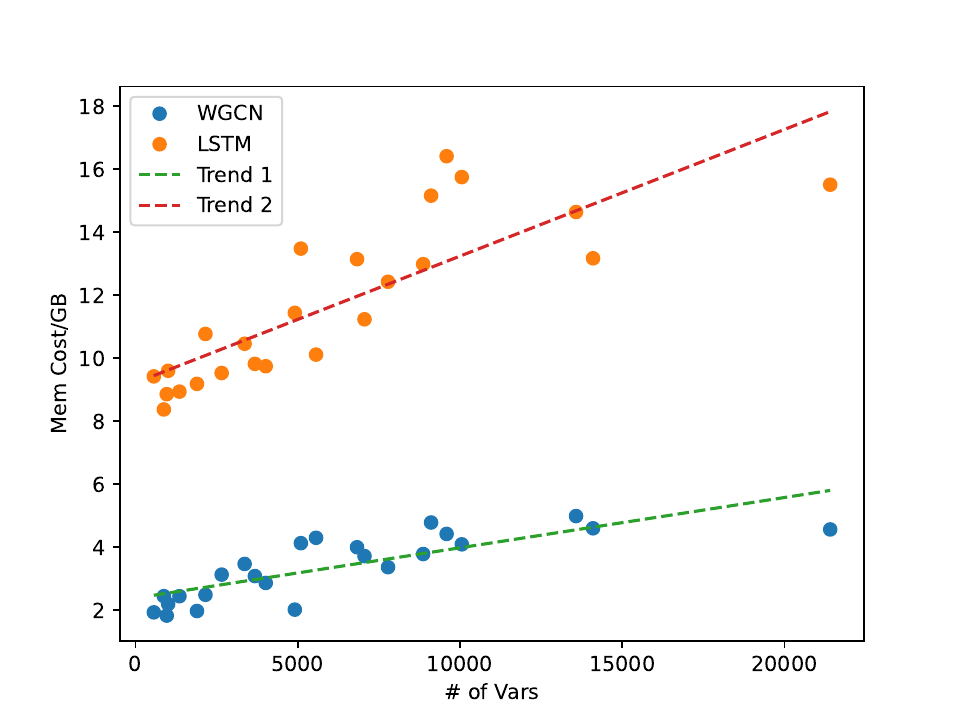}

\label{ablation_fig2a}
\end{minipage}
}
\subfigure[Training time of different neural network]{
\begin{minipage}[t]{0.43\textwidth}
\centering
\includegraphics[width=1\textwidth]{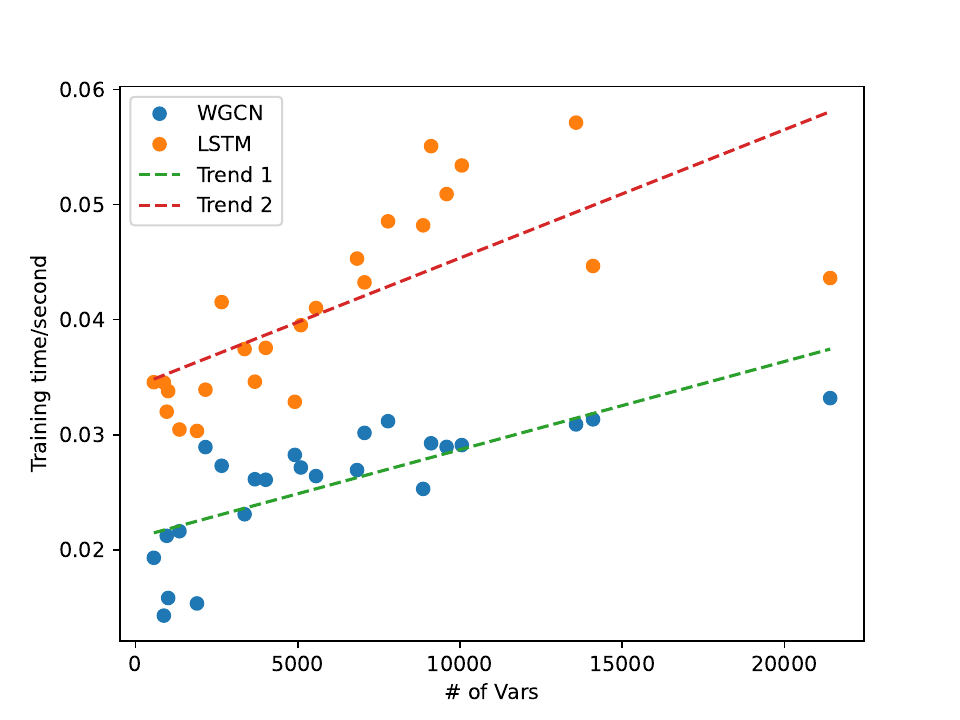}

\label{ablation_fig2b}
\end{minipage}
}
\caption{The scalability: The memory cost (a) and training time (b) according to different number of variables.}
\label{ablation_fig2}
\end{figure}

We plot scatter graphs in Fig. \ref{ablation_fig2}, showing memory consumption and training time of varying sizes of data samples under different computation units. WGCN uses significantly less memory across all sizes compared to LSTM, around only 30\%. This stark difference suggests that model with WGCN is memory-efficient, potentially allowing for large or complex problem-solving without compromising system resources like GPU. WGCN also demonstrates faster training speeds across all data sample sizes. On average, WGCN is found to be approximately 40\% quicker than LSTM. This speed advantage can reduce the time required for training models, thereby increasing the efficiency of model development process.

\section{Conclusion}
In this paper, we presented a novel approach specifically tailored to the UNSAT problem solving in LEC process. Our approach, termed as IB-Net, introduces a GNN model to predict a decision strategy queue and decision scores, serving as a powerful assistant tool for CDCL SAT solvers.

Our comprehensive set of experiments established the superiority of IB-Net in accelerating SAT solving,  significantly outperforming existing methods. This performance gain is particularly pronounced within the context of LEC. Furthermore, our approach has demonstrated the capability to successfully solve instances that were previously deemed unsolvable using traditional SAT solvers. 

The impact of IB-Net extends beyond its immediate performance advantages. By combining machine learning techniques with traditional SAT solvers, the unique blend of these disciplines can enable more efficient and robust solvers. As we continue to refine and expand on our method, we envision a future where the performance gains offered by approaches like IB-Net become standard in the realm of SAT solving.

\clearpage

\end{document}